\documentclass[11pt,a4paper]{article}

\usepackage{url,fullpage,graphicx,subcaption,amssymb,amsmath}
\usepackage[section,above,below]{placeins}
\usepackage{ifpdf}
\ifpdf
\fi
\graphicspath{{expe/}}

\newtheorem{lemma}{Lemma}

\title{Guaranteed bounds on the Kullback-Leibler divergence of univariate mixtures using piecewise log-sum-exp inequalities}

\author{Frank Nielsen\thanks{Frank Nielsen is with \'Ecole Polytechnique
and Sony Computer Science Laboratories Inc.
 E-mail: {\tt Frank.Nielsen@acm.org} } 
\and
Ke Sun \thanks{Ke Sun is with   \'Ecole Polytechnique, E-mail: {\tt sunk.edu@gmail.com}.}
}

\date{}

\begin{document}

\maketitle 
 
\begin{abstract}
Information-theoretic measures such as the entropy, cross-entropy and the Kullback-Leibler divergence between two mixture models is a core primitive in many signal processing tasks.
Since the Kullback-Leibler divergence of mixtures provably does not admit a closed-form formula, 
it is in practice either estimated using costly Monte-Carlo stochastic integration, approximated, or bounded using various techniques.
We present a fast and generic method that builds algorithmically closed-form lower and upper bounds on the entropy, the cross-entropy and the Kullback-Leibler divergence of mixtures.
We illustrate the versatile method by reporting on our experiments for approximating the Kullback-Leibler divergence between univariate exponential mixtures, 
Gaussian mixtures, Rayleigh mixtures, and Gamma mixtures.
\end{abstract}

\def\cH{{H_\times}}
\def\cL{{L_\times}}
\def\cU{{U_\times}}
\def\bbR{\mathbb{R}}
\def\calM{\mathcal{M}}
\def\dx{\mathrm{d}x}
\def\dy{\mathrm{d}y}
\def\dt{\mathrm{d}t}
\def\KL{\mathrm{KL}}
\def\JS{\mathrm{JS}}
\def\lse{\mathrm{lse}}
\def\erf{\mathrm{erf}}
\def\Erf#1{\mathrm{erf}{\left(#1\right)}}
\def\calX{\mathcal{X}}
\def\calE{\mathcal{E}}
\def\calD{\mathcal{D}}

\section{Introduction}  

Mixture models are commonly used in signal processing.
A typical scenario is to use mixture models~\cite{GMM-2008,RMMs-2011,EMMs-2006} to {\em smoothly} model histograms. 
For example, Gaussian Mixture Models (GMMs) can be used to convert grey-valued images into binary images by building a
GMM fitting the image intensity histogram and then choosing the threshold as the average of the Gaussian means~\cite{GMM-2008} to binarize the image. 
Similarly, Rayleigh Mixture Models (RMMs) are often used in ultrasound imagery~\cite{RMMs-2011} to model histograms, and perform segmentation by classification.
When using mixtures, a fundamental primitive is to define a proper {\em statistical distance} between them.
The {\em Kullback-Leibler divergence}~\cite{EIT-2012}, also called  {\em relative entropy}, is the most commonly used distance:
Let $m(x)=\sum_{i=1}^k w_i p_i(x)$ and $m'(x)=\sum_{i=1}^{k'} w_i' p_i'(x)$ be two finite statistical density\footnote{The cumulative density function (CDF) of a mixture is like its density also a convex combinations of the component CDFs.
But beware that a mixture is {\em not} a sum of random variables. Indeed, sums of RVs have convolutional densities.} mixtures of $k$ and $k'$ components, respectively.
In statistics, the mixture components $p_i(x)$ are often parametric: $p_i(x)=p_i(x;\theta_i)$, where $\theta_i$ is a vector of parameters.
For example, a mixture of Gaussians (MoG also used as a shortcut instead of GMM) has its component distributions 
parameterized by its mean $\mu_i$ and its covariance matrix $\Sigma_i$ (so that the parameter vector is $\theta_i=(\mu_i,\Sigma_i)$).
Let $\calX=\{x \in \mathbb{R}\ :\ p(x;\theta)>0 \}$ denote the support of the component distributions, and
denote by $\cH(m,m')=-\int_\calX m(x)\log m'(x) \dx$ the {\em cross-entropy}~\cite{EIT-2012} between mixtures $m$ and $m'$. 
Then the Kullback-Leibler (KL) divergence between   two continuous mixtures of densities $m$ and $m'$ is given by:

\begin{eqnarray}
\KL(m:m') &=&\int_\calX m(x)\log \frac{m(x)}{m'(x)} \dx,\\
 &=&\cH(m,m')-H(m),
\end{eqnarray}

\noindent with $H(m)=\cH(m,m)=-\int m(x)\log m(x)\dx$ denoting the {\em Shannon entropy}~\cite{EIT-2012}.
The notation  ``:'' is used instead of the usual coma ``,'' notation to emphasize that the distance is {\em not} a metric distance 
since it is not symmetric ($\KL(m:m')\not =\KL(m':m)$), and that it further does not satisfy the triangular inequality~\cite{EIT-2012} of metric distances
 ($\KL(m:m')+\KL(m':m'') \not\geq \KL(m:m'')$).
When the natural base of the logarithm is chosen, we get a differential entropy measure expressed in {\em nat} units.
Alternatively, we can also use the base-2 logarithm ($\log_2 x =\frac{\log x}{\log 2}$) and get the entropy expressed in {\em bit} units. 
Although the KL divergence is available in closed-form for many distributions 
(in particular as equivalent Bregman divergences for exponential families~\cite{BD-2005}), it was proven
that the Kullback-Leibler divergence between two (univariate) GMMs is {\em not analytic}~\cite{KLnotanalytic-2004} (the particular case of mixed-Gaussian of two components with same variance was analyzed in~\cite{mixedGaussian2compo-2008}). 
See appendix \ref{sec:KLnonanalytic} for an analysis.
Note that the differential entropy may be negative: For example, the differential entropy of a univariate Gaussian distribution is 
$\log(\sigma\sqrt{2\pi e})$, and is therefore negative when the standard variance $\sigma<\frac{1}{\sqrt{2\pi e}}$ (about $0.242$).
We consider continuous distributions with entropies well-defined (entropy may be undefined for singular distributions like Cantor's distribution).

Thus many approximations techniques have been designed to beat the computational-costly Monte-Carlo (MC) stochastic {\em estimation}:
$\widehat{\KL_s}(m:m') = \frac{1}{s} \sum_i \log\frac{m(x_i)}{m'(x_i)}$ with $x_1, \ldots, x_s\sim m(x)$ 
($s$ independently and identically distributed (iid) samples $x_1, \ldots, x_s$).
The MC estimator is asymptotically consistent, $\lim_{s\rightarrow\infty} \widehat{\KL_s}(m:m') = \KL(m:m')$, so that the ``true value'' of the KL of mixtures
is estimated in practice by taking a very large sampling (say, $s=10^9$). 
However, we point out that the MC estimator is a stochastic approximation, and therefore {\em does not} guarantee deterministic bounds (confidence intervals may be used).  
Deterministic lower and upper bounds of the integral can be obtained by various numerical integration techniques using quadrature rules.
We refer to~\cite{HGMM-2008,GM-KLIEP,KLGMM-2012,comix-2016} for the current state-of-the-art approximation techniques and bounds on the KL of GMMs.
The latest work for computing the entropy of GMMs is~\cite{HGMM-2016}: It considers arbitrary finely tuned bounds of computing the entropy of {\em isotropic Gaussian} mixtures (a case encountered when dealing with KDEs, kernel density estimators).
However, there is catch in the technique of~\cite{HGMM-2016}: It relies on solving for the unique roots of some log-sum-exp equations (See Theorem~1 of~\cite{HGMM-2016}, pp. 3342) that do not admit a closed-form solution.
Thus it is hybrid method that contrasts with our combinatorial approach

In information geometry~\cite{IG-2016}, a {\em mixture family} of linearly independent probability distributions $p_1(x), ..., p_{k}(x)$ is defined by the convex combination of those non-parametric component distributions.
$m(x;\eta)=\sum_{i=1}^k \eta_i p_i(x)$. 
A mixture family induces a dually flat space where the Kullback-Leibler divergence is equivalent to a Bregman divergence~\cite{BD-2005,IG-2016} defined on the $\eta$-parameters.
However, in that case, the Bregman convex generator $F(\eta)=\int m(x;\eta)\log m(x;\eta)\dx$ (the Shannon information) is
 {\em not} available in closed-form for mixtures.
Except for the family of multinomial distribution that is both a mixture family (with closed-form $\KL(m:m')=\sum_{i=1}^k m_i\log\frac{m_i}{m_i'}$, the discrete KL~\cite{EIT-2012}) and an exponential family~\cite{IG-2016}.

In this work, we present a simple and efficient intuitive method that builds algorithmically a closed-form formula that guarantees both deterministic lower and upper bounds   on the KL divergence  within an {\em additive factor} 
of $\log k+\log k'$. We then further refine our technique to get improved adaptive bounds.
For univariate GMMs, we get the non-adaptive additive bounds in $O(k\log k+k'\log k')$ time, and the adaptive bounds in $O(k^2+k'^2)$ time.

To illustrate our generic technique, we demonstrate it  on finding bounds for the KL of Exponential Mixture Models (EMMs), Rayleigh mixtures, Gamma mixtures and GMMs.

The paper is organized as follows:
Section~\ref{sec:alg} describes the algorithmic construction of the formula using piecewise log-sum-exp inequalities for the cross-entropy and the Kullback-Leibler divergence.
Section~\ref{sec:ent} instantiates this algorithmic principle to the entropy and discusses several related work.
Section~\ref{sec:exp}  reports on experiments on several   mixture families.
Finally, Section~\ref{sec:concl} concludes this work by discussing extensions to other statistical distances.
Appendix~\ref{sec:KLnonanalytic} proves that the Kullback-Leibler divergence of mixture models is not analytic.
Appendix~\ref{sec:appendix} reports the closed-form formula for the KL divergence between scaled and truncated distributions of the same exponential family~\cite{ef-flashcard-2009} (that includes Rayleigh, Gaussian and Gamma distributions among others).

\section{A generic combinatorial bounding algorithm \label{sec:alg}}

Let us bound the cross-entropy $\cH(m:m')$ by deterministic lower and upper bounds, $\cL(m:m') \leq \cH(m:m')\leq \cU(m:m')$,
so that the bounds on the  Kullback-Leibler divergence $\KL(m:m')=\cH(m:m')-\cH(m:m)$ follows as:

\begin{equation} 
\cL(m:m')-\cU(m:m) \leq \KL(m:m')
\leq \cU(m:m')-\cL(m:m).
\end{equation}

Since the cross-entropy of two mixtures $\sum_{i=1}^k w_ip_i(x)$ and $\sum_{j=1}^{k'} w_j'p_j'(x)$:
\begin{equation} 
\cH(m:m')= -\int_\calX \left(\sum_i w_ip_i(x)\right)\log \left(\sum_j w_j'p_j'(x)\right)\dx
\end{equation}
 has a log-sum term of positive arguments, we shall use bounds on the log-sum-exp (lse) function~\cite{lse-2014,boosting-2010}:
$$
\lse(\{x_i\}_{i=1}^l) =\log\left(\sum_{i=1}^l e^{x_i}\right).
$$
We have the following inequalities:

\begin{equation} 
\max\{x_i\}_{i=1}^l < \lse(\{x_i\}_{i=1}^l)\leq \log l + \max \{x_i\}_{i=1}^l.
\end{equation}
The left-hand-side (lhs) inequality holds because $\sum_{i=1}^l e^{x_i}>\max \{e^{x_i}\}_{i=1}^l =\exp(\max \{x_i\}_{i=1}^l)$ since $e^x>0, \forall x\in\mathbb{R}$, and
the right-hand-side (rhs) inequality follows from the fact that  $\sum_{i=1}^l e^{x_i}\leq l \max  \{e^{x_i}\}_{i=1}^l = l\exp(\max \{x_i\}_{i=1}^l)$.
The lse function is convex (but not strictly convex) and enjoys the following translation identity property: $\lse(\{x_i\}_{i=1}^l)=c+\lse(\{x_i-c\}_{i=1}^l)$, $\forall{c}\in\mathbb{R}$.
Similarly, we can also lower bound the lse function by $\log l + \min \{x_i\}_{i=1}^l$.
Note that we could write {\em equivalently}  that for $l$ {\em positive} numbers  $x_1,\ldots, x_l$, we have:
\begin{equation}
\max\left\{\log \max\{x_i\}_{i=1}^l,\log l+ \log \min \{x_i\}_{i=1}^l\right\} 
\le \log \sum_{i=1}^l x_i
\le \log l + \log \max \{x_i\}_{i=1}^l.
\end{equation}

Therefore we shall bound the integral term $\int_\calX m(x) \log \left(\sum_{j=1}^{k'} w_j'p_j'(x)\right) \dx$ using piecewise lse inequalities where the min/max are kept unchanged.

Using the log-sum-exp inequalities, we get
$\cL(m:m') =A(m:m') -\log k'$ and
$\cU(m:m')=A(m:m')$ where

\begin{equation} 
\boxed{A(m:m') = -\int_\calX m(x) \max \{\log w_j' + \log p'_j(x)\}_{j=1}^{k'} \dx.}
\end{equation}

In order to calculate  $A(m:m')$ efficiently using closed-form formula, let us compute the {\em upper envelope} of the $k'$ 
real-valued functions $\calE(x)=\max\{w_j'p_j'(x)\}_{j=1}^{k'}$ defined on the support $\calX$.
This upper envelope can be computed exactly using techniques of computational geometry~\cite{CG-2000,Env3D-2010} 
provided that we can calculate the roots of the equality equation of weighted components: $w_r'p_r'(x)=w_s'p_s'(x)$.
(Although this amounts to solve quadratic equations for Gaussian or Rayleigh distributions, the roots may not
 always be available in closed form, say for example for Weibull distributions.)

\begin{figure}[t]
\centering
\includegraphics[width=.8\textwidth]{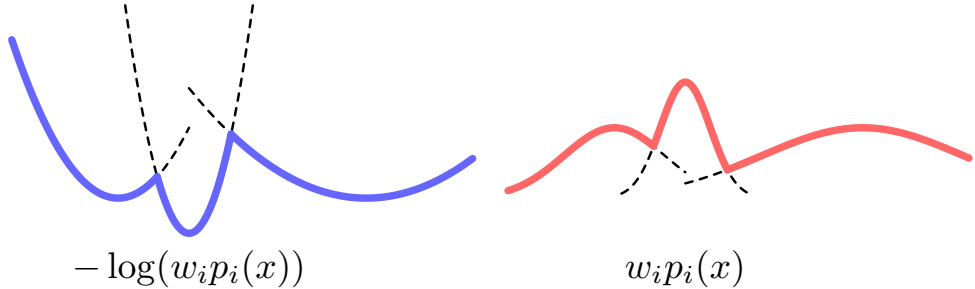}
\caption{Lower envelope of parabolas corresponding to the
upper envelope of weighted components of a Gaussian mixture (here, $k=3$ components).
\label{fig:envelope}}
\end{figure}

Let the upper envelope be combinatorially described by $m$ {\em elementary interval pieces}
 defined on support intervals $I_r=(a_r,a_{r+1})$ partitioning the support $\calX=\uplus_{r=1}^m I_r$ (with $a_1=\min \calX$ and $a_{m+1}=\max \calX$).
Observe that on each interval $I_r$, the maximum of the functions $\max\{w_j'p_j'(x)\}_{j=1}^{k'}$ is given by $w_{\delta(r)}'p_{\delta(r)}'(x)$, where
$\delta(r)$ indicates the weighted component dominating all the others (that is, the arg max of $\{w_j'p_j'(x)\}_{j=1}^{k'}$ for any $x\in I_r$).
To fix ideas, when mixture components are univariate Gaussians, the upper envelope $\calE(x)$ amounts
to find equivalently the {\em  lower envelope} of $k'$ parabola (see Fig.~\ref{fig:envelope}) which has linear complexity, and can be computed in $O(k'\log k')$-time~\cite{lowerenv-1995}, or
in output-sensitive time $O(k'\log m)$~\cite{env-1998}, where $m$ denotes the number of parabola segments of the envelope.
When the Gaussian mixture components have all the same weight and variance (e.g., kernel density estimators), 
the upper envelope amounts to find a lower envelope of cones: $\min_{j}\vert{}x-\mu'_{j}\vert$ (a Voronoi diagram in arbitrary dimension).

To proceed once the envelope has been built, we need to calculate two types of {\em definite integrals} on those elementary intervals: 
(i)  the {\em probability mass} in an
interval  $\int_a^b p(x)\dx=\Phi(b)-\Phi(a)$ where $\Phi$ denotes the 
Cumulative Distribution Function (CDF), and
(ii) the {\em partial cross-entropy} $-\int_a^b p(x)\log p'(x)\dx$~\cite{cHEF-2010}.
Thus let us define these two quantities:
\begin{eqnarray} 
C_{i,j}(a,b)&=&-\int_a^b w_i p_i(x) \log (w_j'p_j'(x))   \dx, \\
M_i(a,b) &=& -\int_a^b w_i p_i(x) \dx .
\end{eqnarray}

Then we get the term $A(m:m')$ as
$$
A(m:m') =
\sum_{r=1}^m \sum_{s=1}^k
\left(
M_s(a_r,a_{r+1})\log w_{\delta(r)}+C_{s,\delta(r)}(a_r,a_{r+1})
\right).
$$

The size of the lower/upper bound formula depends on the complexity of the upper envelope, and of the closed-form expressions of the integral terms 
$C_{i,j}$ and $M_i(a,b)$.
In general, when weighted component densities intersect in at most $p$ points, the
complexity is related to the Davenport-Schinzel sequences~\cite{DS-1995}.
It is quasi-linear for bounded $p=O(1)$, see~\cite{DS-1995}. 

Note that in symbolic computing, the  Risch semi-algorithm~\cite{symbint-2005} solves the problem of computing indefinite integration 
in terms of elementary functions provided that there exists an oracle (hence the term semi-algorithm) for checking whether an expression is equivalent to zero or not
 (however it is unknown whether there exists an algorithm implementing the oracle or not).

We presented the technique by bounding the cross-entropy (and entropy) to deliver lower/uppers bounds on the KL divergence.
When only the KL divergence needs to be bounded, we rather consider the ratio term $\frac{m(x)}{m'(x)}$. 
This requires to partition the support $\calX$ into elementary intervals by
overlaying the critical points of both the lower and upper envelopes of $m(x)$ and $m'(x)$.
In a given elementary interval, since $\max(k \min_i\{w_i p_i(x)\}, \max_i\{w_i p_i(x)\}) \leq m(x)\leq k \max_i\{w_i p_i(x)\}$ , we then consider the inequalities: 
\begin{eqnarray}
\frac{\max(k \min_i\{w_i p_i(x)\}, \max_i\{w_i p_i(x)\})}{k \max_j\{w_j' p_j'(x)\}}  \leq \frac{m(x)}{m'(x)}
\leq \frac{k \max_i\{w_i p_i(x)\}}{\max(k  \min_j\{w_j' p_j'(x),\max_j\{w_j' p_j'(x)\}\})}.
\end{eqnarray}
We now need to compute definite integrals of the form $\int_a^b w_1 p(x;\theta_1)\log\frac{w_2 p(x;\theta_2)}{w_3 p(x;\theta_3)}\dx$ (see Appendix~\ref{sec:appendix} for explicit formulas when considering scaled and truncated exponential families~\cite{ef-flashcard-2009}).
(Thus for exponential families, the ratio of densities remove the auxiliary carrier measure term.)%

We call these bounds CELB and CEUB that stands for Combinatorial Envelope Lower and Upper Bounds, respectively.

\subsection{Tighter adaptive bounds\label{sec:adapbound}}

We shall now consider {\em data-dependent} bounds improving   over the additive $\log k+\log k'$ non-adaptive bounds.
Let $t_i(x_1, \ldots, x_k)=\log (\sum_{j=1}^k e^{x_j-x_i})$. 
Then $\lse(x_1, \ldots, x_k) =x_i+t_i(x_1, \ldots, x_k)$
for all $i\in [k]$. We denote by $x_{(1)}, \ldots, x_{(k)}$ the  sequence of numbers sorted in increasing order.

Clearly, when $x_i=x_{(k)}$ is chosen as the maximum element, we have 
$$
\log \left(\sum_{j=1}^k e^{x_j-x_i}\right)= \log \left(1+ \sum_{j=1}^{k-1} e^{x_{j}-x_{(k)}}\right)\leq \log k
$$
since $x_{j}-x_{(k)}\leq 0$ for all $j\in [k]$.

Also since $e^{x_j-x_i}=1$ when $j=i$ and $e^{x_j-x_i}>0$, we have necessarily $t_i(x_1, \ldots, x_k)>0$ for any $i\in [k]$.
Since it is an identity for all $i\in [k]$, we minimize $t_i(x_1, \ldots, x_k)$ by maximizing $x_i$, and therefore,
we have $\lse(x_1, \ldots, x_k)\leq x_{(k)}+ t_{(k)}(x_1, \ldots, x_k)$ where $t_{(k)}$ yields the smallest residual.

When considering 1D GMMs, let us now bound $t_{(k)}(x_1, \ldots, x_k)$ in a combinatorial range $I_s=(a_{s},a_{s+1})$ of the lower envelope of parabolas.
Let $m=\delta(s)$ denote the index of the dominating weighted component in the range:

$$
\forall x\in I_s,\forall i \quad
\exp\left(-\log\sigma_i-\frac{(x-\mu_i)^2}{2\sigma_i^2}+\log w_i\right)
\leq
\exp\left(-\log\sigma_m-\frac{(x-\mu_m)^2}{2\sigma_m^2}+\log w_m\right).
$$

Thus we have:
\begin{eqnarray*}
\lefteqn{\log m(x) = \log \frac{w_m}{\sigma_m \sqrt{2\pi}} -\frac{(x-\mu_m)^2}{2\sigma_m^2} }\\
&&+ \log \left(1+\sum_{j\not =m} \exp\left(-\frac{(x-\mu_i)^2}{2\sigma_i^2}+\log\frac{w_i}{\sigma_i}
+\frac{(x-\mu_m)^2}{2\sigma_m^2}-\log\frac{w_m}{\sigma_m}\right)\right)
\end{eqnarray*}

Now consider the ratio term:
$$
r_{i,m}(x)=\exp\left(-\frac{(x-\mu_i)^2}{2\sigma_i^2}
+\log \frac{w_i\sigma_m}{w_m\sigma_i}+\frac{(x-\mu_m)^2}{2\sigma_m^2}\right)
$$

It is maximized in $I_s=(a_s,a_{s+1})$ by maximizing equivalently the following quadratic equation:

$$
l_{i,m}(x) = -\frac{(x-\mu_i)^2}{2\sigma_i^2}
+\log \frac{w_i\sigma_m}{w_m\sigma_i}+\frac{(x-\mu_m)^2}{2\sigma_m^2}
$$

Setting the derivative to zero ($l_{i,m}'(x)=0$), we get the root (when $\sigma_i\not =\sigma_m$)

$$
x_{i,m} = \frac{\frac{\mu_m}{\sigma_m^2} - \frac{\mu_i}{\sigma_i^2}}{\frac{1}{\sigma_m^2} - \frac{1}{\sigma_i^2} }.
$$

If $x_{i,m}\in I_s$ then we report the value $R_{i,m}(s)=
\max\{r_{i,m}(x_{i,m}),r_{i,m}(a_s),r_{i,m}(a_{s+1})\}$, otherwise the maximum
value of $r_{i,m}(x)$ in the slab $I_s$ is obtained by considering
$R_{i,m}(s)=\max\{r_{i,m}(a_s),r_{i,m}(a_{s+1})\}$.

Then $t_m$ is bounded in range $I_s$ by:
$$
t_m = \log \left(1+\sum_{i\not =m} R_{i,m}(s)\right)\leq \log k
$$

In practice, we always get better bounds using the data-dependent technique at the expense of 
computing  overall the $O(k^2)$ intersection points of the pairwise densities.

We call those bounds CEALB and CEAUB for Combinatorial Envelope Adaptive Lower Bound (CEALB) and Combinatorial Envelope Adaptive Upper Bound (CEAUB).
 
Let us illustrate one scenario where this adaptive technique yields very good approximations:
For example, consider a GMM with all variance $\sigma^2$ tending to zero (a mixture of $k$ Diracs).
Then in a combinatorial slab, we have $R_{i,m}(s)\rightarrow 0$ for all $i\not =m$, and we get tight bounds.

Notice that we could have also upper bounded $\int_{a_s}^{a_{s+1}} \log m(x) \dx$ by $(a_{s+1}-a_{s})m(a_{s},a_{s+1})$ where $m(x,x')$ denotes
the maximal value of the mixture density in the range $(a_{i},a_{i+1})$.
The maximal value is either found at the slab extremities, 
or is a mode of the GMM: It then requires to find  the modes of a GMM~\cite{modefinding-2000}, for which no analytical solution is known in general.

\subsection{Another derivation using the  arithmetic-geometric mean inequality}

Let us start by considering  the inequality of arithmetic and geometric weighted means applied to the mixture component distributions:

$$
m'(x)=\sum_i w_i' p(x;\theta_i') \geq \prod_i p(x;\theta_i')^{w_i'} 
$$
with equality iff.   $p(x;\theta_1')= \ldots =p(x;\theta_k')$.

To get a tractable formula with a positive remainder of the log-sum term $\log m'(x)$, 
we need to have the log argument greater or equal to $1$, and thus we shall write the positive remainder:

$$
R(x)=\log\left(\frac{m'(x)}{\prod_i p(x;\theta_i')^{w_i'} }\right) \geq 0.
$$

Therefore, we can decompose the log-sum into a tractable part $\log\prod_i p(x;\theta_i')^{w_i'}$ and the remainder as:

$$
\log m'(x)= \sum_i w_i' \log p(x;\theta_i') + \log \left( \frac{m'(x)}{\prod_i p(x;\theta_i')^{w_i'} }\right).
$$

Clearly, since the geometric mean is a  quasi-arithmetic mean, it ranges between the extrema values of its elements:

$$
\min_i p(x;\theta_i') \leq \prod_i p(x;\theta_i')^{w_i'}\leq \max_i p(x;\theta_i').
$$

Therefore, we have the following inequalities for the remainder:

$$
\log\frac{m'(x)}{\max_i p(x;\theta_i')}
\leq \log \frac{m'(x)}{\prod_i p(x;\theta_i')^{w_i'}}
\leq \log \frac{m'(x)}{\min_i p(x;\theta_i')}. 
$$

And we proceed similarly by computing the lower and upper envelope, and bounding $m'(x)$ by $\min_i w_i'p_i'(x) \leq m'(x) \leq \max_i w_i'p_i'(x)$.

Next, we instantiate this method for the prominent cases of exponential mixture models, Gaussian mixture models and Rayleigh mixture models often used to model intensity histograms in image~\cite{GMM-2008} and ultra-sound~\cite{RMMs-2011} processing, respectively.

\subsection{Case studies}
 
\subsubsection{The case of exponential mixture models}

An exponential distribution has density $p(x;\lambda)=\lambda \exp(-\lambda x)$
defined on $\calX=[0,\infty)$ for $\lambda>0$.  Its CDF is $\Phi(x;\lambda)=1-
\exp(-\lambda x)$.  Any two components $w_1p(x;\lambda_1)$ and
$w_2p(x;\lambda_2)$ (with $\lambda_1\neq\lambda_2$) have a unique
intersection point
\begin{equation}
x^\star=\frac{\log(w_2\lambda_2)-\log(w_1\lambda_1)}{\lambda_2-\lambda_1}
\end{equation}
if $x^\star\ge0$; otherwise they do not intersect.
The basic quantities to evaluate the bounds are
\begin{align}
C_{i,j}(a,b)
=
&
\log\left( \lambda'_jw'_j \right) M_i(a,b)
+w_i\lambda'_j\left[
\left(a+\frac{1}{\lambda_i}\right) e^{-\lambda_ia}
-
\left(b+\frac{1}{\lambda_i}\right) e^{-\lambda_ib}
\right],\\
M_i(a,b) =&
-w_i \left( e^{-\lambda_ia} - e^{-\lambda_ib} \right).
\end{align}

\subsubsection{The case of Rayleigh mixture models}

A Rayleigh distribution has density
$p(x;\sigma)=\frac{x}{\sigma^2}\exp\left(-\frac{x^2}{2\sigma^2}\right)$, defined
on  $\calX=[0,\infty)$ for $\sigma>0$. Its CDF is
$\Phi(x;\sigma)=1-\exp\left(-\frac{x^2}{2\sigma^2}\right)$. Any two components
$w_1p(x;\sigma_1)$ and $w_2p(x;\sigma_2)$ (with $\sigma_1\neq\sigma_2$)
must intersect at $x_0=0$ and can have at most one other intersection point 
\begin{equation}
x^\star=\sqrt{
\log\frac{w_1\sigma_2^2}{w_2\sigma_1^2}
/\left( \frac{1}{2\sigma_1^2}- \frac{1}{2\sigma_2^2} \right)}
\end{equation}
if the square root is well defined and $x^\star>0$. We have
\begin{align}
C_{i,j}(a,b) =&
\log\frac{{w}'_j}{(\sigma_j')^2} M_i(a,b)
+\frac{w_i}{2(\sigma_j')^2}
\left[
(a^2+2\sigma_i^2)e^{-\frac{a^2}{2\sigma_i^2}}
-
(b^2+2\sigma_i^2)e^{-\frac{b^2}{2\sigma_i^2}}
\right]\nonumber\\
&-w_i \int_{a}^b 
\frac{x}{\sigma_i^2}\exp\left(-\frac{x^2}{2\sigma_i^2}\right)
\log{x} \dx,\label{eq:numint}\\
M_i(a,b) =&
-w_i \left( e^{-\frac{a^2}{2\sigma_i^2}} - e^{-\frac{b^2}{2\sigma_i^2}} \right).
\end{align}
The last term in Eq.~(\ref{eq:numint}) does not have a simple closed form (it requires the exponential integral Ei).  
One need a
numerical integrator to compute it.

\subsubsection{The case of Gaussian mixture models}

The Gaussian density
$p(x;\mu,\sigma)=\frac{1}{\sqrt{2\pi}\sigma}e^{-{(x-\mu)^2}/(2\sigma^2)}$
has support $\calX=\mathbb{R}$ and parameters $\mu\in\mathbb{R}$ and $\sigma>0$. Its CDF is
$\Phi(x;\mu,\sigma)=\frac{1}{2}\left[1+\erf(\frac{x-\mu}{\sqrt{2}\sigma})
\right]$, where $\erf$ is the Gauss error function. The intersection point
$x^\star$ of two components $w_1p(x;\mu_1,\sigma_1)$ and
$w_2p(x;\mu_2,\sigma_2)$ can be obtained by solving the quadratic equation
$\log\left(w_1p(x^\star;\mu_1,\sigma_1)\right)=\log\left(w_2p(x^\star;\mu_2,\sigma_2)\right)$,
which gives at most two solutions. As shown in Fig.~(\ref{fig:envelope}), the
upper envelope of Gaussian densities correspond to the lower envelope of
parabolas.  We have
\begin{align}
C_{i,j}(a,b) =&
M_i(a,b)
\left(
\log{w}_j' - \log\sigma_j'-\frac{1}{2}\log(2\pi)
-\frac{1}{2(\sigma_j')^2}
\left((\mu_j' - \mu_i)^2+\sigma_i^2\right)
\right) \nonumber\\
&+ \frac{w_i\sigma_i}{2\sqrt{2\pi}(\sigma_j')^2}
\left[
(a+\mu_i-2\mu_j') e^{-\frac{(a-\mu_i)^2}{2\sigma_i^2}}
-
(b+\mu_i-2\mu_j') e^{-\frac{(b-\mu_i)^2}{2\sigma_i^2}}
\right],\\
M_i(a,b) =& -\frac{w_i}{2}
\left(
\erf\left(\frac{b-\mu_i}{\sqrt{2}\sigma_i}\right)
-
\erf\left(\frac{a-\mu_i}{\sqrt{2}\sigma_i}\right)
\right).
\end{align}

\subsubsection{The case of gamma distributions}

For simplicity, we only consider $\gamma$-distributions with fixed shape parameter $k>0$ and varying scale $\lambda>0$.
The density is defined on $(0,\infty)$ as $p(x;k,\lambda)=\frac{x^{k-1} e^{-\frac{x}{\lambda}}}{\lambda^k \Gamma(k)}$,
where $\Gamma(\cdot)$ is the gamma function.
Its CDF is $\Phi(x;k,\lambda) = \gamma(k,x/\lambda)/\Gamma(k)$, where
$\gamma(\cdot,\cdot)$ is the lower incomplete gamma function.
Two weighted gamma densities $w_1p(x;k,\lambda_1)$ and $w_2p(x;k,\lambda_2)$ (with $\lambda_1\neq\lambda_2$)
intersect at a unique point 
\begin{equation}
x^\star
=
\frac{ \log\frac{w_1}{\lambda_1^k} - \log\frac{w_2}{\lambda_2^k} }
{\frac{1}{\lambda_1}- \frac{1}{\lambda_2}}
\end{equation}
if $x^\star>0$; otherwise they do not intersect. From straightforward
derivations,
\begin{align}
C_{i,j}(a,b) &= \log\frac{w'_j}{(\lambda'_j)^k\Gamma(k)}M_i(a,b)
+w_i \int_a^b
\frac{x^{k-1} e^{-\frac{x}{\lambda_i}}}{\lambda_i^k \Gamma(k)}
\left( \frac{x}{\lambda'_j} - (k-1)\log{x} \right) dx, \label{eq:gamma1}\\
M_i(a,b)     &= -\frac{w_i}{\Gamma(k)} 
\left( \gamma\left(k,\frac{b}{\lambda_i}\right)
     - \gamma\left(k,\frac{a}{\lambda_i}\right) \right).
\end{align}
Again, the last term in Eq.~(\ref{eq:gamma1}) relies on numerical integration.

\section{Upper-bounding the differential entropy of a mixture\label{sec:ent}}

First, consider a finite parametric mixture $m(x)=\sum_{i=1}^k w_i p_i(x;\theta_i)$.
Using the chain rule of the entropy, we end up with the well-known lemma:

\begin{lemma}
The entropy of a mixture is upper bounded by the sum of the entropy of its marginal mixtures:
$H(m)\le\sum_{i=1}^k H(m_i)$, where $m_i$ is the 1D marginal mixture with respect to variable $x_i$.
\end{lemma}

Since the  1D marginals of a multivariate GMM  are univariate GMMs, we thus get (loose) upper bound.
In general, a generic  sample-based
probabilistic bounds is reported for the entropies of distributions
with given support~\cite{Hprobaupperbound-2008}: The method considers the empirical cumulative distribution function from a iid finite sample set of size $n$ to build probabilistic upper and lower piecewisely linear CDFs given a deviation probability threshold. 
It then builds algorithmically between those two bounds the maximum entropy distribution~\cite{Hprobaupperbound-2008} with a so-called string-tightening algorithm.

Instead, proceed as follows: 
Consider finite mixtures of component distributions defined on the full support $\mathbb{R}^d$ that have finite component means and variances (like exponential families). Then we shall use the fact that the maximum differential entropy with prescribed mean and variance is a Gaussian distribution\footnote{In general, the maximum entropy with moment constraints yields as a solution an exponential family.}, and conclude the upper bound by plugging the mixture mean and variance in the differential entropy formula of the Gaussian distribution.

Wlog, consider GMMs $m(x)=\sum_{i=1}^k w_i p(x;\mu_i,\Sigma_i)$ ($\Sigma_i=\sigma_i^2$ for univariate Gaussians).
The mean $\bar{\mu}$ of the mixture is $\bar{\mu}=\sum_{i=1}^k w_i \mu_i$ and the variance is
$\bar{\sigma}^2=E[m^2]-E[m]^2$.
Since $E[m^2]=\sum_{i=1}^k w_i \int x^2 p(x;\mu_i,\Sigma_i)\dx$
and $\int x^2 p(x;\mu_i,\Sigma_i)\dx = \mu_i^2+\sigma_i^2$, we deduce that
$$
\bar{\sigma}^2
= \sum_{i=1}^k w_i (\mu_i^2+\sigma_i^2) - \left(\sum_{i=1}^k w_i \mu_i\right)^2
= \sum_{i=1}^k w_i \left[ (\mu_i-\bar\mu)^2 + \sigma_i^2 \right].
$$

The entropy of a random variable with a prescribed variance $\bar{\sigma}^2$ is maximal for the Gaussian distribution with the same variance $\bar{\sigma}^2$, see~\cite{EIT-2012}.
Since the differential entropy of a Gaussian is $\log (\bar{\sigma}\sqrt{2\pi e})$, we deduce that the entropy of the GMM is upper bounded by
$$
H(m) \leq 
  \frac{1}{2} \log (2\pi e)
+ \frac{1}{2} \log \sum_{i=1}^k w_i \left[(\mu_i-\bar\mu)^2+\sigma_i^2\right].
$$

This upper bound generalizes to arbitrary dimension, we get the following lemma:

\begin{lemma}
The entropy of a $d$-variate GMM $m(x)=\sum_{i=1}^k w_i p(x;\mu_i,\Sigma_i)$ is upper bounded by
$\frac{d}{2}\log(2\pi e)+\frac{1}{2}\log \det \Sigma$, where
$\Sigma= \sum_{i=1}^k w_i (\mu_i\mu_i^\top+\Sigma_i)
- \left(\sum_{i=1}^k w_i \mu_i\right) \left(\sum_{i=1}^k w_i \mu_i^\top\right)$.
\end{lemma}

In general, exponential families have finite moments of any order~\cite{ef-flashcard-2009}: 
In particular, we have $E[t(X)]=\nabla F(\theta)$ and $V[t(X)]=\nabla^2 F(\theta)$.
For the Gaussian distribution, we have the sufficient statistics $t(x)=(x,x^2)$
so that $E[t(X)]=\nabla F(\theta)$ yields the mean and variance from the log-normalizer.

Note that this bound (called the Maximum Entropy Upper Bound in~\cite{HGMM-2016}, MEUB) is tight when the GMM approximates a single Gaussian.
It is fast to compute compared to the bound reported in~\cite{HGMM-2008} that uses Taylor' s expansion of the log-sum of the mixture density.

A similar argument cannot be applied for a lower bound since a GMM with a given variance may have 
entropy tending to $-\infty$ as follows: 
Wlog., assume  the $2$-component mixture's mean is zero, and that the variance equals $1$ by taking
$m(x) =  \frac{1}{2} G(x;-1,\epsilon) + \frac{1}{2} G(x;1,\epsilon)$ where $G$ denotes the Gaussian density.
Letting $\epsilon\rightarrow 0$, we get the entropy tending to $-\infty$.

We remark that our log-exp-sum inequality technique yields a $\log 2$ additive approximation range for the case of a Gaussian mixture
with two components. It thus generalizes the bounds reported in~\cite{mixedGaussian2compo-2008} to arbitrary variance mixed Gaussians. 

Let $U(m:m')$ and $L(m:m')$ denotes the deterministic upper and lower bounds,
and  $\Delta(m:m')=U(m:m')-L(m:m')\geq 0$ denotes the bound gap where 
the true value of the KL divergence belongs to.
In practice, we seek matching lower and upper bounds that minimize the 
bound gap.

Consider the lse inequality $\log k+\min_i x_i\leq \lse(x_1, \ldots, 
x_k)\leq \log k+\max_i x_i$.
The gap of that ham-sandwich inequality is $\max_i x_i-\min_i x_i$ since 
the $\log k$ terms cancel out.
This gap improves over the $\log k$ gap of $\max_i x_i\leq \lse(x_1, 
\ldots, x_k)\leq \log k+\max_i x_i$
when $\max_i x_i-\min_i x_i\leq \log k$.

For log-sum terms of mixtures, we have $x_i=\log p_i(x)+\log w_i$.
$$
\max_i x_i-\min_i x_i = \log \frac{\exp(\max_i x_i)}{\exp(\min_i x_i )}
$$

For the differential entropy, we thus have
$$
-\sum_s \int_{I_s} m(x)\log \max_i w_ip_i(x) \dx
\le H(m) \le
-\sum_s \int_{I_s} m(x)\log \min_i w_ip_i(x) \dx
$$

Therefore the gap is:
$$
\Delta=\sum_s \int_{I_s} m(x)\log \frac{\max_i w_ip_i(x)}{\min_i w_ip_i(x) } \dx
$$

Thus to compute the gap error bound of the differential entropy,
we need to integrate terms
$$
\int w_ap(x;\theta_a) \log  \frac{w_bp_b(x)}{w_cp_c(x)} \dx
$$

See appendix~\ref{sec:appendix} for a closed-form formula when dealing with
exponential family components.

\section{Experiments\label{sec:exp}}

We perform an empirical study to verify our theoretical bounds.  We 
simulate four pairs of mixture models $\{(\mathtt{EMM}_1, \mathtt{EMM}_2),
(\mathtt{RMM}_1,  \mathtt{RMM}_2), (\mathtt{GMM}_1,  \mathtt{GMM}_2),
(\mathtt{GaMM}_1, \mathtt{GaMM}_2) \}$ as the test subjects. 
The component type is implied by the model name. The components of each mixture
model are given as follows.
\begin{enumerate}
\item $\mathtt{EMM}_1$'s components, in the form $(\lambda_i,w_i)$, are given by
$(0.1,1/3)$, $(0.5,1/3)$, $(1,1/3)$; $\mathtt{EMM}_2$'s components are $(2,
0.2)$, $(10,0.4)$, $(20,0.4)$.

\item $\mathtt{RMM}_1$'s components, in the form $(\sigma_i,w_i)$, are given by
$(0.5,1/3)$, $(2,1/3)$, $(10,1/3)$; $\mathtt{RMM}_2$ consists of $(5, 0.25)$,
$(60,0.25)$, $(100,0.5)$.

\item $\mathtt{GMM}_1$'s components, in the form $(\mu_i,\sigma_i,w_i)$, are
$(-5, 1, 0.05)$, $( -2, 0.5, 0.1)$, $( 5, 0.3, 0.2)$, $( 10, 0.5, 0.2 )$, $( 15,
0.4, 0.05 )$ $(25, 0.5, 0.3)$, $( 30, 2, 0.1 )$; $\mathtt{GMM}_2$ consists of
$(-16, 0.5, 0.1)$, $(-12, 0.2, 0.1)$, $(-8, 0.5, 0.1)$, $(-4, 0.2, 0.1)$,  $( 0,
0.5, 0.2 )$, $(4, 0.2, 0.1)$, $( 8, 0.5, 0.1)$, $(12, 0.2, 0.1)$), $(16, 0.5,
0.1 )$. 

\item $\mathtt{GaMM}_1$'s components, in the form $(k_i,\lambda_i,w_i)$, are
$(2,0.5,1/3)$, $(2,2,1/3)$, $(2,4,1/3)$; $\mathtt{GaMM}_2$ consists of 
$(4,5,1/3)$, $(4,8,1/3)$, $(4,10,1/3)$.
\end{enumerate}

We compare the proposed bounds with Monte-Carlo estimation with different sample
sizes in the range $\{10, 10^2, 10^3, 10^4\}$. For each sample size
configuration, Monte-Carlo estimation is repeated for 100 times to get the
statistics.  Fig.~(\ref{fig:results})(a-d) shows the input signals as well as
the estimation results, where the proposed bounds CELB, CEUB, CEALB, CEAUB are
presented as horizontal lines, and the Monto-Carlo estimations over different
sample sizes are presented as error bars.  We can loosely consider the average
Monte-Carlo output with the largest sample size ($10^4$) as the underlying
truth, which is clearly inside our bounds.  This serves as an empirical
justification on the correctness of the bounds.

A key observation is that the bounds can be \emph{very tight}, especially when
the underlying KL divergence has a large magnitude, e.g.
$\KL(\mathtt{RMM}_2:\mathtt{RMM}_1)$.  This is because the gap between the lower
and upper bounds is always guaranteed to be within $\log{k}+\log{k}'$.  Because
KL is unbounded measure~\cite{EIT-2012}, in the general case two mixture models
may have a large KL. Then our approximation gap is relatively very small.  On
the other hand, we also observed that the bounds in certain cases, e.g.
$\KL(\mathtt{EMM}_2:\mathtt{EMM}_1)$, are not as tight as the other cases. When
the underlying KL is small, the bound is not as informative as the general case. 

Comparatively, there is a significant improvement of the data-dependent bounds
(CEALB and CEAUB) over the combinatorial bounds (CELB and CEUB). In all
investigated cases, the adaptive bounds can roughly shrink the gap by half
of its original size at the cost of additional computation.

Note that, the bounds are accurate and must contain the true value. Monte-Carlo
estimation gives no guarantee on where the true value is. For example, in 
estimating $\KL(\mathtt{GMM}_1:\mathtt{GMM}_2)$, Monte-Carlo estimation based on
$10^3$ samples can go beyond our bounds! It therefore suffers from a larger
estimation error.

We simulates a set of Gaussian mixture models besides the above $\mathtt{GMM}_1$
and $\mathtt{GMM}_2$. Fig.~\ref{fig:hgmm} shows the GMM densities as well as
their differential entropy. A detailed explanation of the components of each GMM
model is omitted for brevity. 

The key observation is that CEUB (CEAUB) is \emph{very tight} in most of the
investigated cases. This is because that the upper envelope that is used to
compute CEUB (CEAUB) gives a very good estimation of the input signal. 

Notice that MEUB only gives an upper bound of the differential entropy as
discussed in section~\ref{sec:ent}. In general the proposed bounds are tighter
than MEUB. However, this is not the case when the mixture components are merged
together and approximate one single Gaussian (and therefore its entropy can be
well apporiximated by the Gaussian entropy), as shown in the last line of
Fig.~\ref{fig:hgmm}.

\begin{figure}[t]
\begin{subfigure}{\textwidth}
\centering
\includegraphics[width=.8\textwidth]{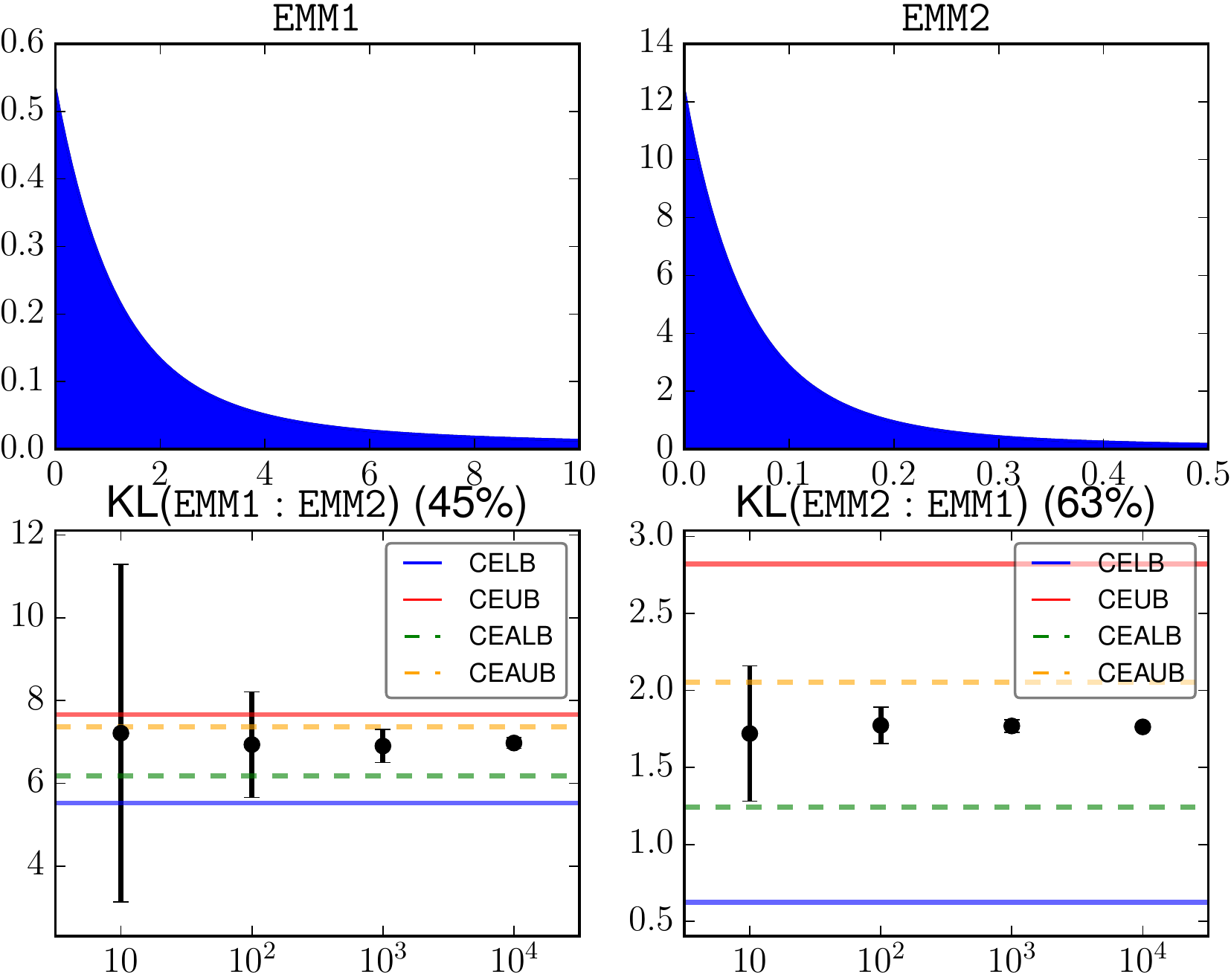}
\caption{KL divergence between two exponential mixture models}
\end{subfigure}
\begin{subfigure}{\textwidth}
\centering
\includegraphics[width=.8\textwidth]{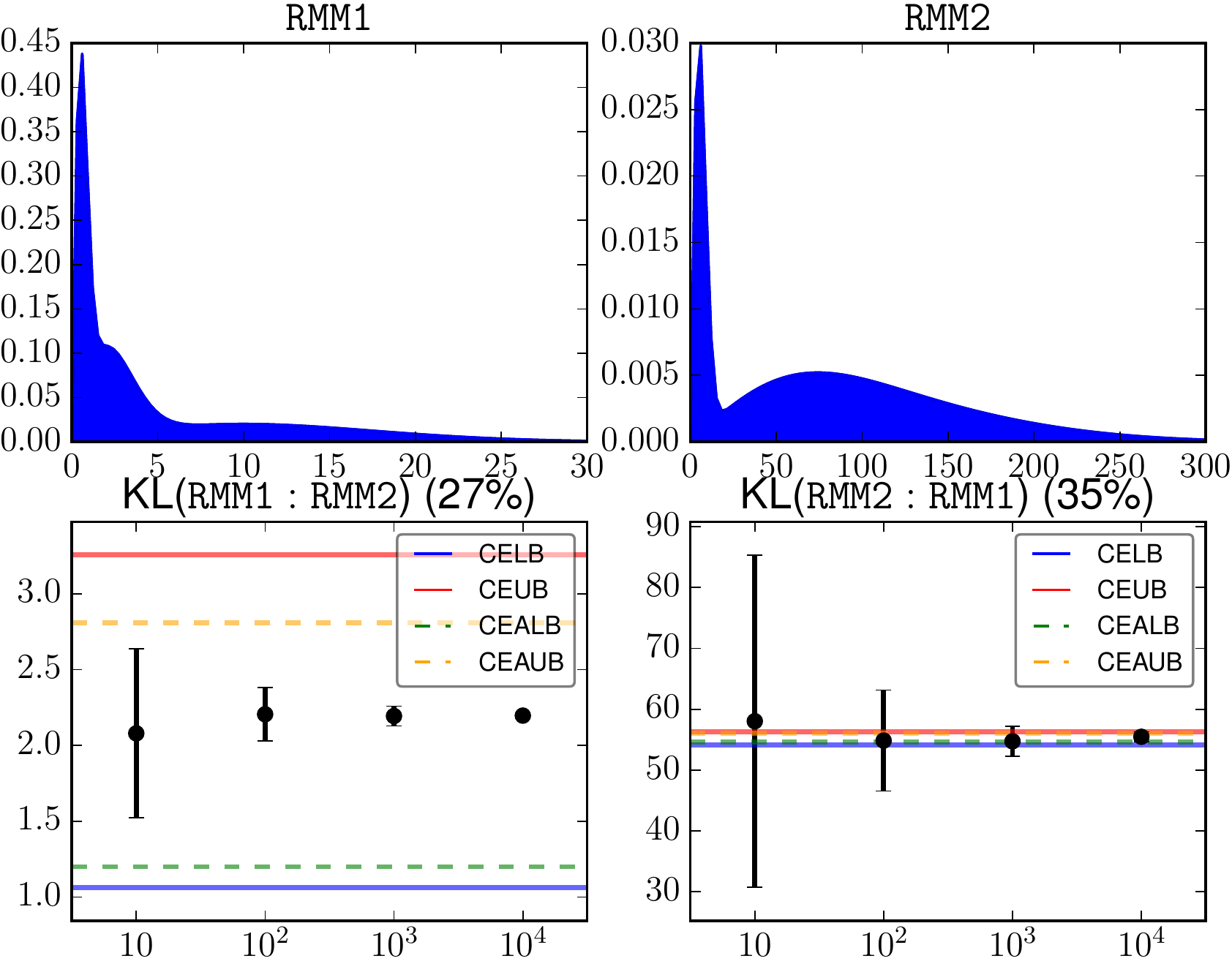}
\caption{KL divergence between two Rayleigh mixture models}
\end{subfigure}
\phantomcaption
\end{figure}

\begin{figure}[t]
\ContinuedFloat 
\begin{subfigure}{\textwidth}
\centering
\includegraphics[width=.8\textwidth]{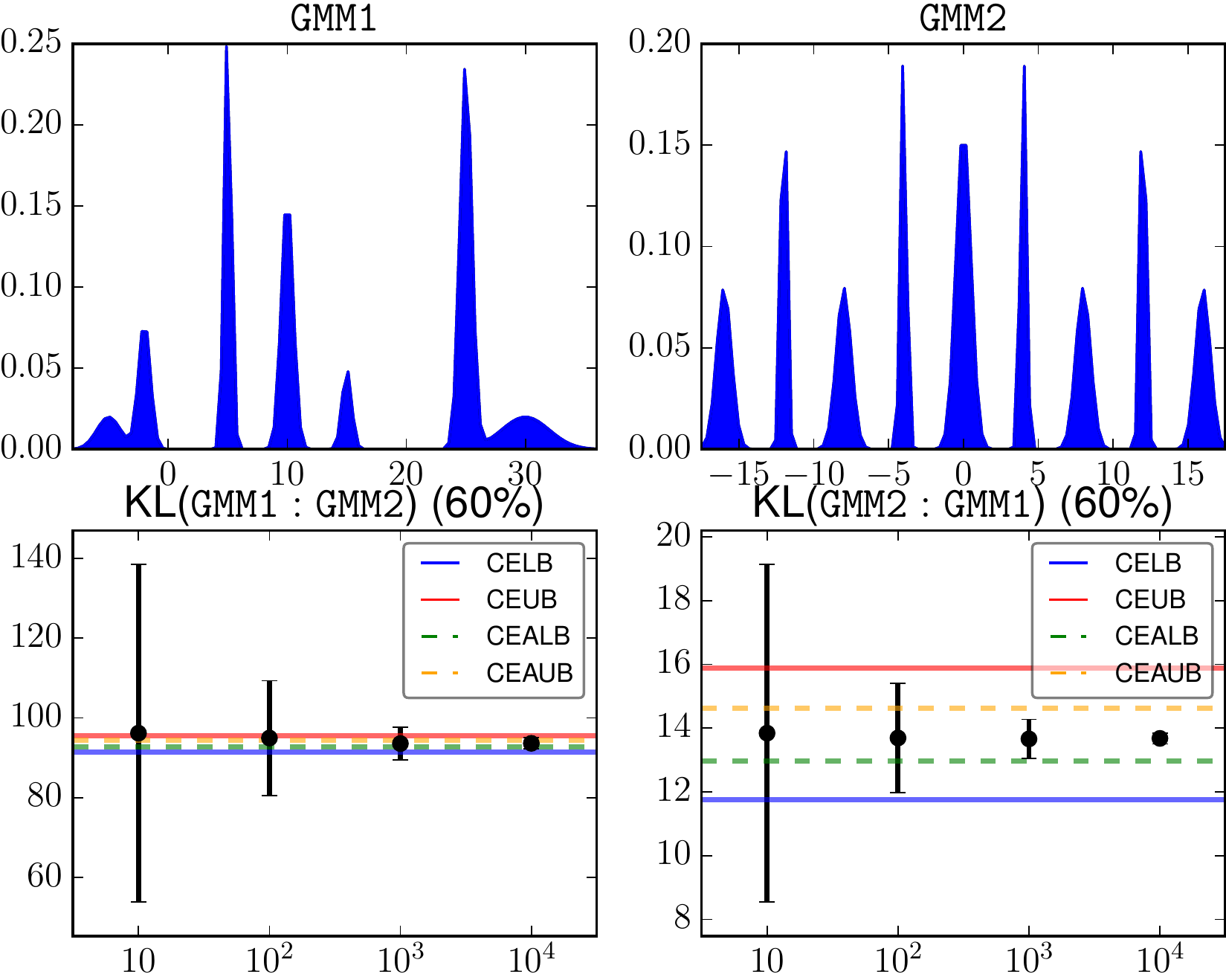}
\caption{KL divergence between two Gaussian mixture models}
\end{subfigure}
\begin{subfigure}{\textwidth}
\centering
\includegraphics[width=.8\textwidth]{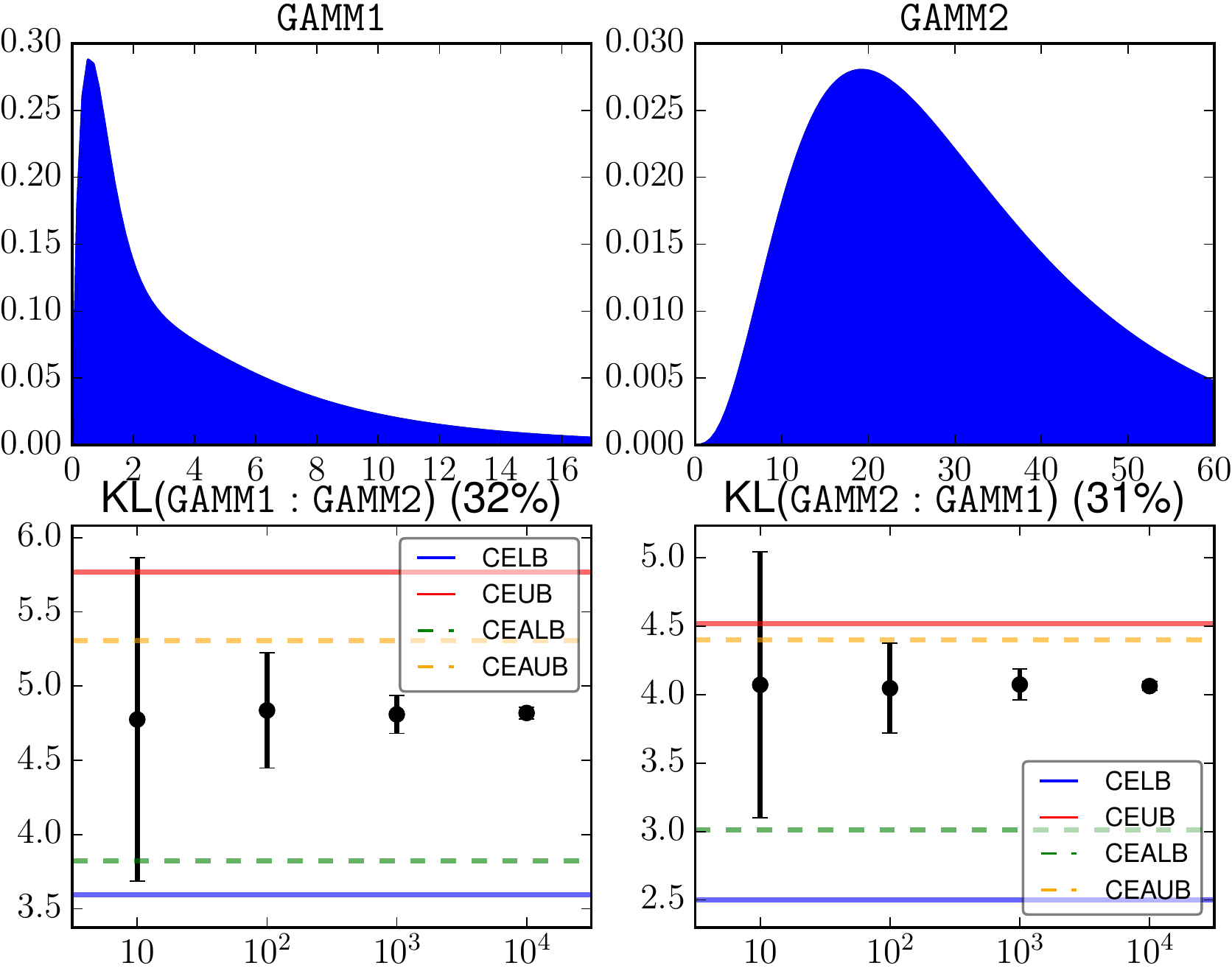}
\caption{KL divergence between two Gamma mixture models}
\end{subfigure}
\caption{Lower and upper bounds on the KL divergence between mixture models.
The percentage in the title indicates the improvement of adaptive bounds
(dashed lines) over the combinatorial bounds (solid lines). The error-bars
show the means and standard deviations of KL (y-axis) by Monte-Carlo estimation
using the corresponding sample size (x-axis).}\label{fig:results}
\end{figure}

\begin{figure}[!t]
\centering
\begin{tabular}{cc}
\includegraphics[width=.49\textwidth]{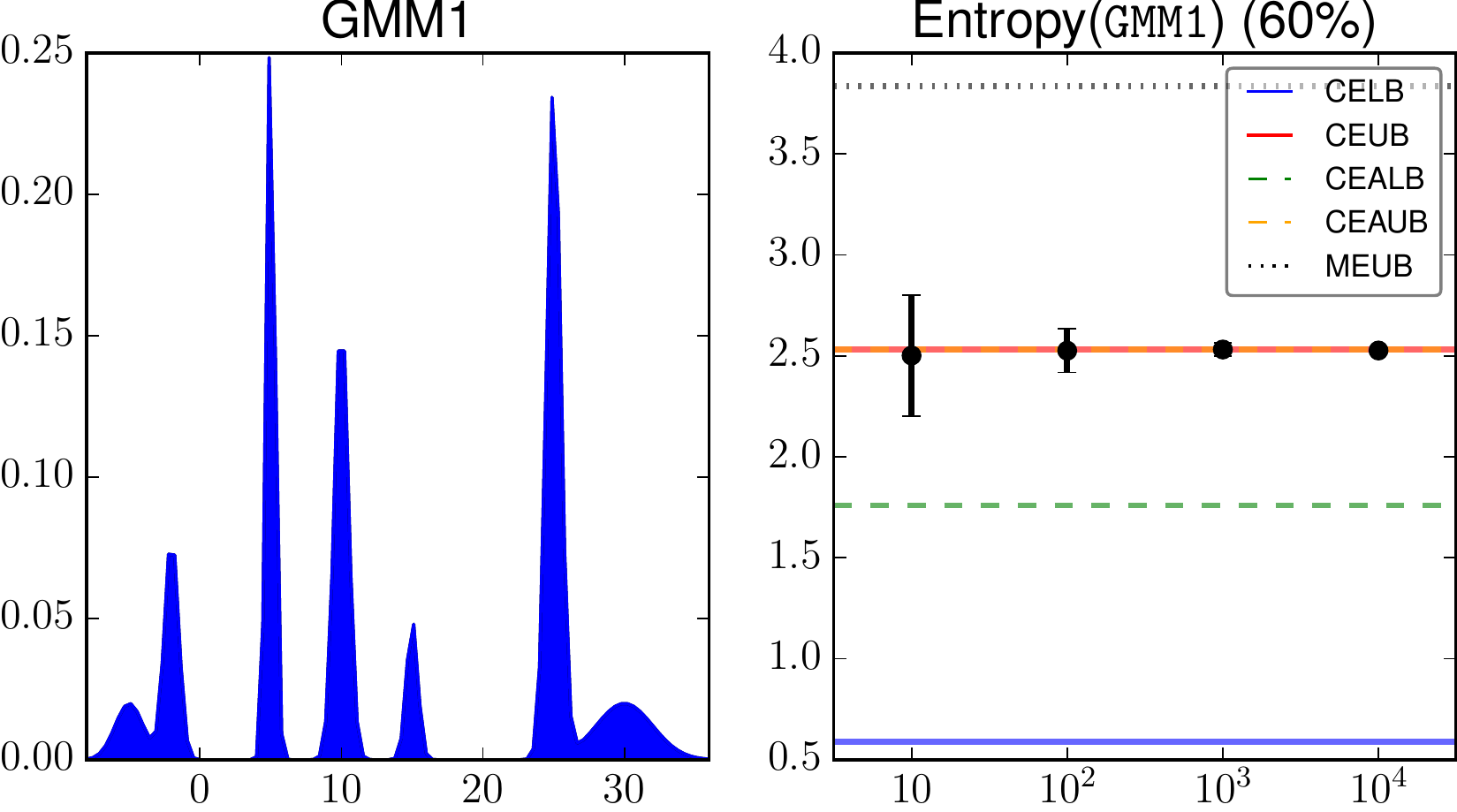}&
\includegraphics[width=.49\textwidth]{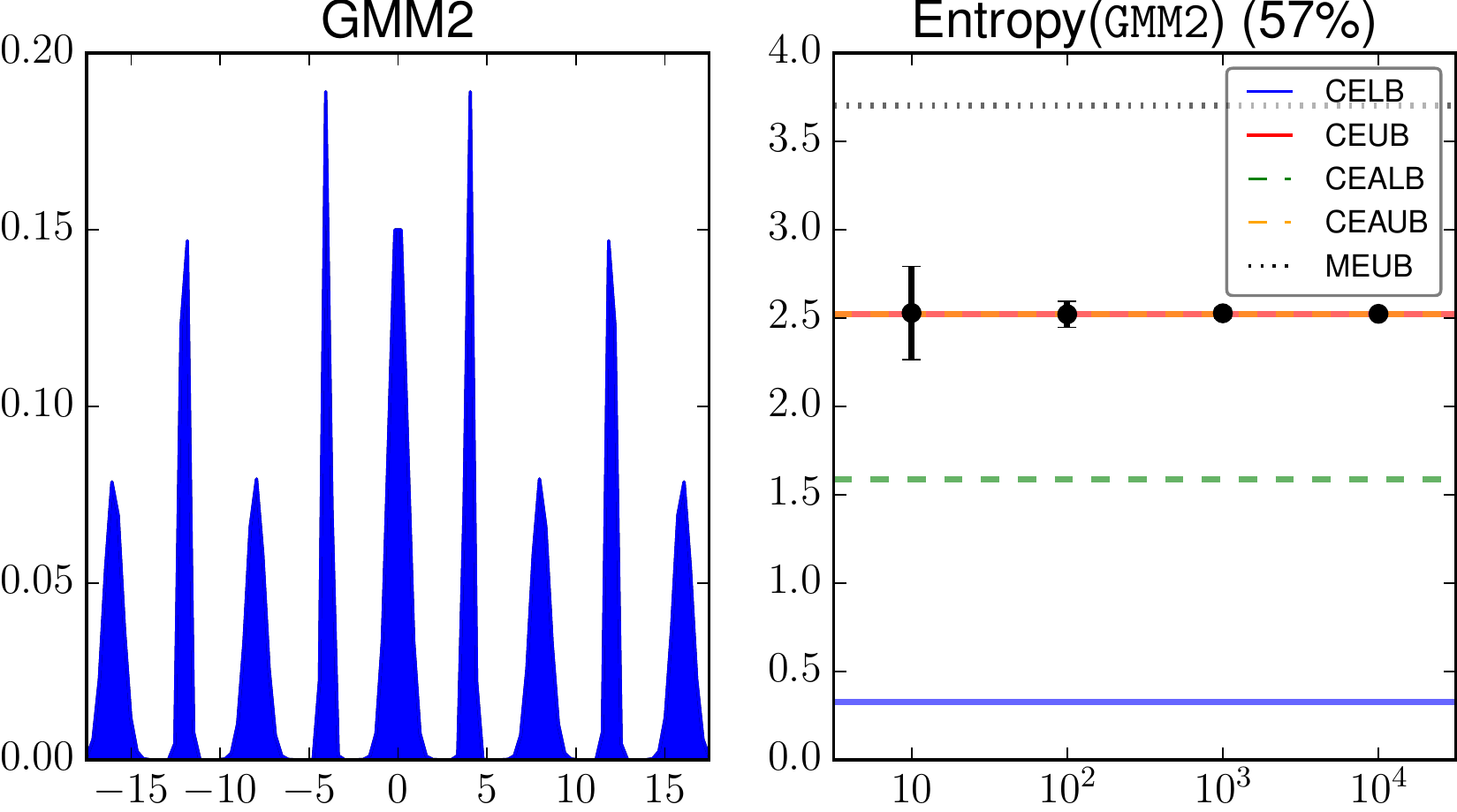}\\
\includegraphics[width=.49\textwidth]{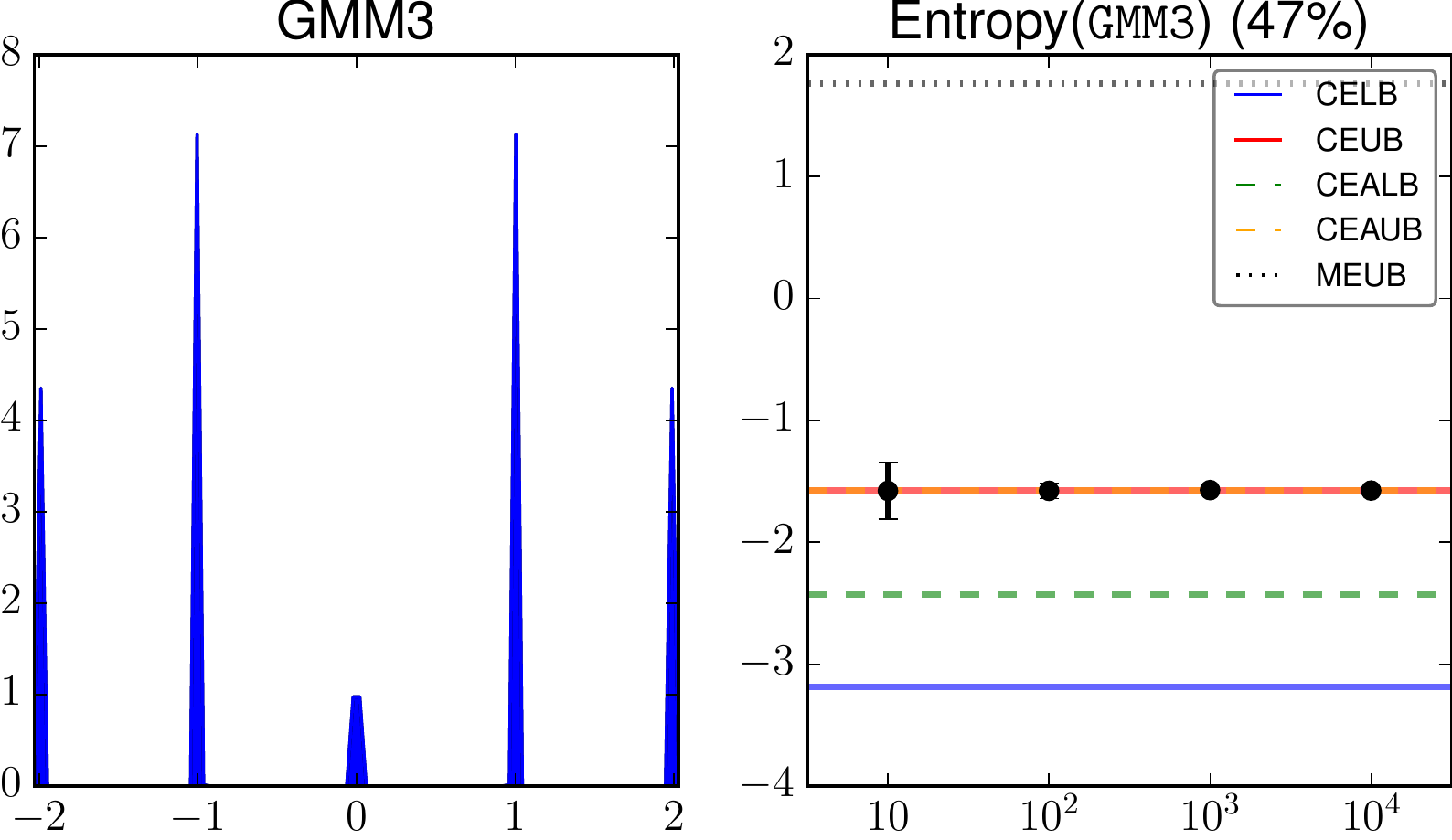}&
\includegraphics[width=.49\textwidth]{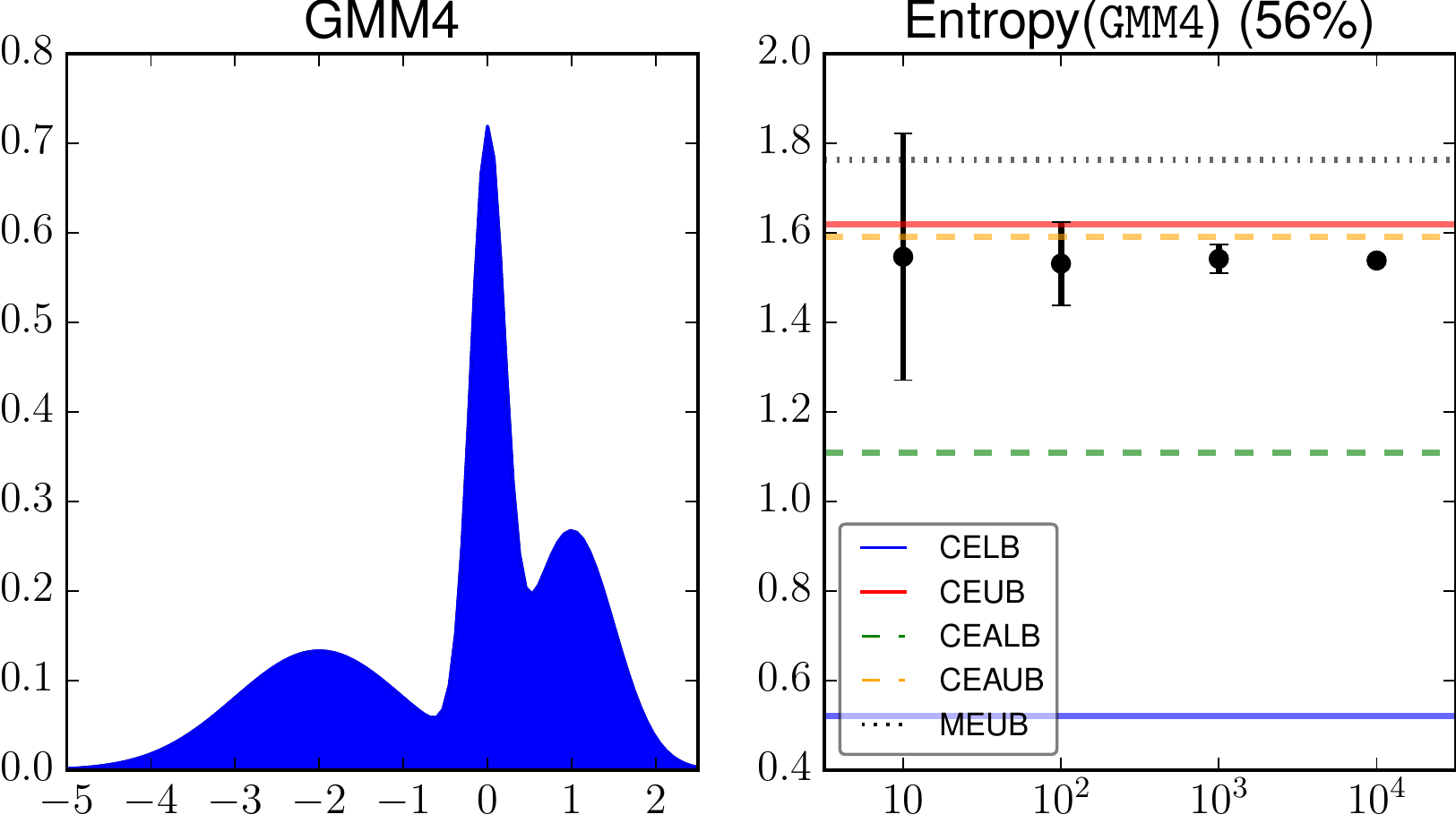}\\
\includegraphics[width=.49\textwidth]{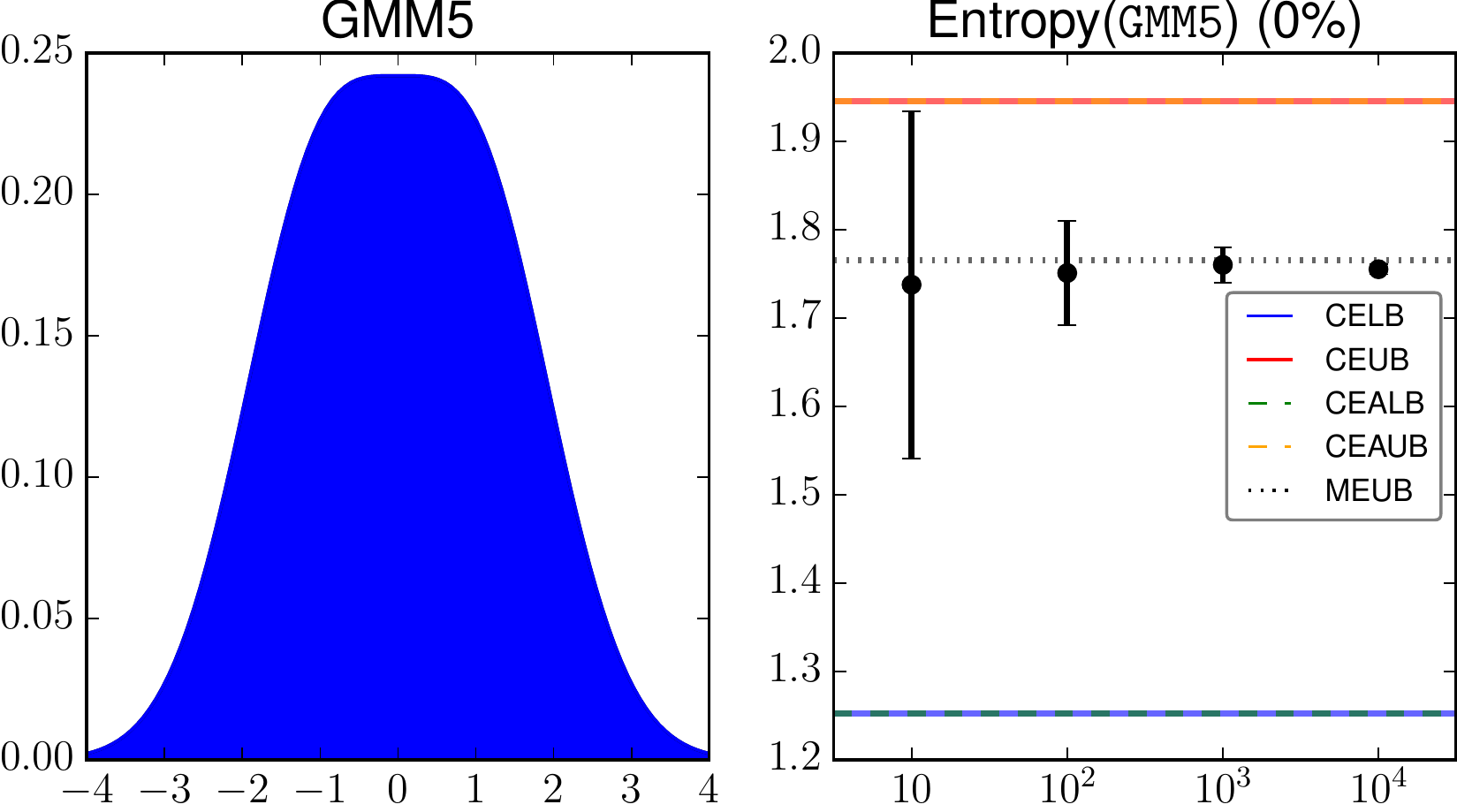}&
\includegraphics[width=.49\textwidth]{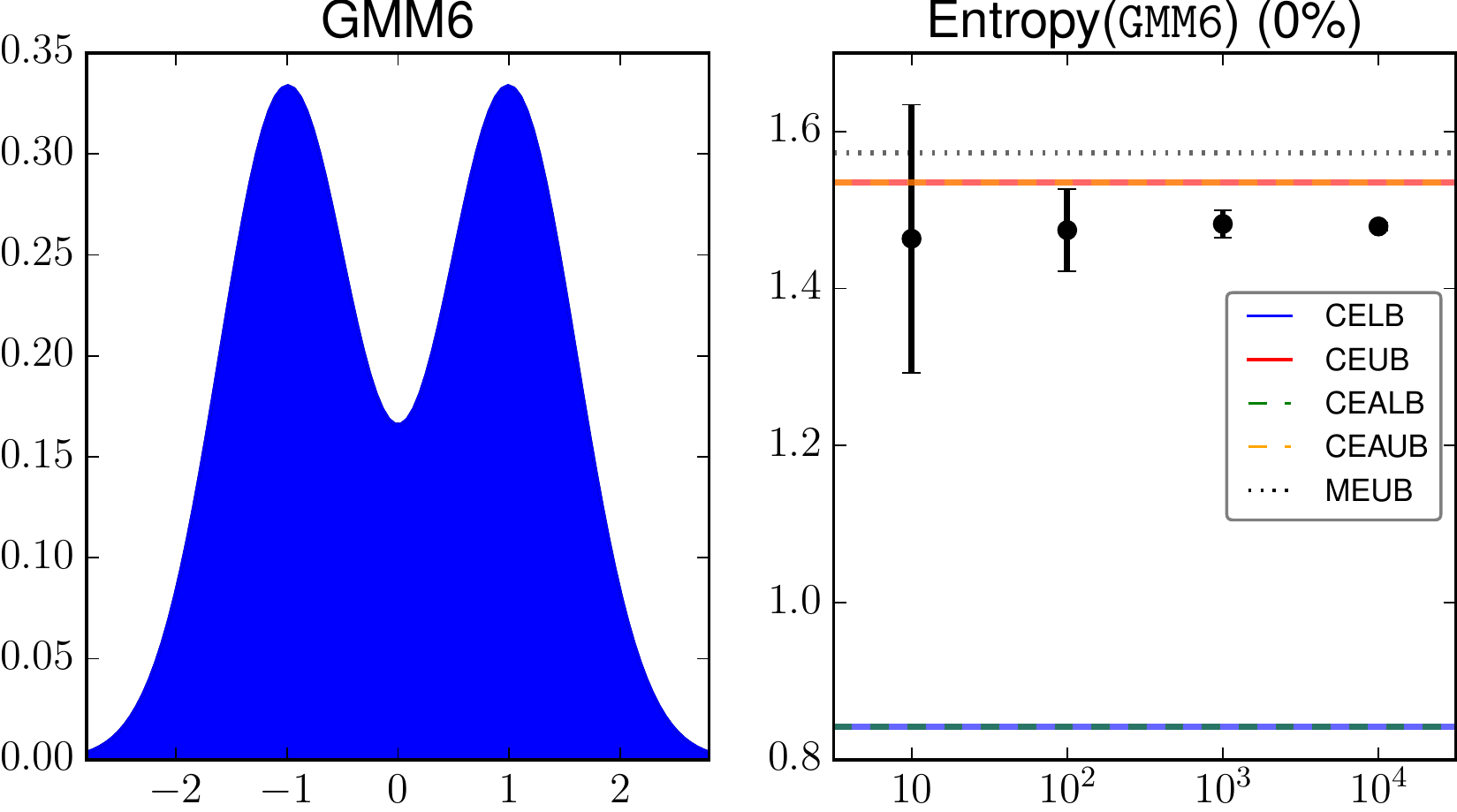}
\end{tabular}
\caption{Lower and upper bounds on the differential entropy of Gaussian mixture models.
On the left of each subfigure is the simulated GMM signal.
On the right of each subfigure is the estimation of its differential entropy.
Compared to Fig.~\ref{fig:results}, the Maximum Entropy Upper Bound shown
by the black dotted line is added.}\label{fig:hgmm}
\end{figure}
 
\section{Concluding remarks and perspectives\label{sec:concl}}

We have presented a fast versatile method to compute bounds on the Kullback-Leibler divergence between mixtures by building algorithmically
 formula. We reported on our experiments for various mixture models in the exponential family.
For univariate GMMs, we get a guaranteed bound of the KL divergence of two mixtures $m$ and $m'$ with $k$ and $k'$ components within an {\em additive} approximation factor
 of $\log k+\log k'$  in $O\left((k+k')\log(k+k')\right)$-time.
Therefore the larger the KL divergence the better the bound when considering a multiplicative $(1+\alpha)$-approximation factor since $\alpha=\frac{\log k+\log k'}{\KL(m:m')}$.
The adaptive bounds is guaranteed to yield better bounds at the expense of computing potentially $O\left(k^2+(k')^2\right)$ intersection points of pairwise weighted components.

Our technique also yields bound for the Jeffreys divergence (the symmetrized KL divergence: $J(m,m')=\KL(m:m')+\KL(m':m)$) and the Jensen-Shannon divergence~\cite{JS-1991} (JS):

$$
\JS(m,m')=
\frac{1}{2}\left(\KL\left(m:\frac{m+m'}{2}\right)+\KL\left(m':\frac{m+m'}{2}\right)\right),
$$
since $\frac{m+m'}{2}$ is a mixture model with $k+k'$ components.
One advantage of this statistical distance is that it is symmetric, always bounded by $\log 2$, and its square root yields a metric distance~\cite{JSmetric-2003}.
The log-sum-exp inequalities may also used to compute some R\'enyi divergences~\cite{Renyi-2011}:
$$
R_\alpha(m,p)  = \frac{1}{\alpha-1} \log \left(\int m(x)^\alpha p(x)^{1-\alpha}\right) \dx,
$$
when $\alpha$ is an integer, $m(x)$ a mixture and $p(x)$ a single (component) distribution.
Getting fast guaranteed tight bounds on statistical distances between mixtures opens many avenues.
For example, we may consider building hierarchical mixture models by merging iteratively two mixture components so that those pair of components is chosen so that  the KL distance between the full mixture and the simplified mixture is minimized.

In order to be useful, our technique is unfortunately limited to univariate mixtures: Indeed, in higher dimensions,
we can still compute the maximization diagram of weighted components (an additively weighted Bregman Voronoi diagram~\cite{BVD-2007,BVD-2010} for components belonging to the same exponential family). However, it becomes more complex to compute in the elementary Voronoi cells $V$, the functions $C_{i,j}(V)$ and $M_i(V)$
 (in 1D, the Voronoi cells are segments). We may obtain hybrid algorithms by approximating or estimating these functions.
In 2D, it is thus possible to obtain lower and upper bounds on the Mutual Information~\cite{LBMI-2011} (MI) when the joint distribution $m(x,y)$ is a 2D mixture of Gaussians:
$$
I(M;M') = \int m(x,y)\log \frac{m(x,y)}{m(x)m'(y)} \dx\dy.
$$
Indeed, the marginal distributions $m(x)$ and $m'(y)$ are univariate Gaussian mixtures.

A Python code implementing those computational-geometric methods for reproducible research is available online at:\\
\begin{center}
\url{https://www.lix.polytechnique.fr/~nielsen/KLGMM/}
\end{center}

Let us now conclude this work by noticing that the Kullback-Leibler of two smooth mixtures can be arbitrarily finely approximated
by a Bregman divergence~\cite{BD-2005}. We loosely derive this observation using two different approaches:

\begin{itemize}

\item Continuous mixture distributions have smooth densities that can be arbitrarily closely approximated using
a {\em single distribution} (potentially multi-modal) belonging to the
 Polynomial Exponential Families~\cite{cobb-1983,pef-2016} (PEFs).
A polynomial exponential family of order $D$ has log-likelihood $l(x;\theta)\propto \sum_{i=1}^D\theta_i x^i$: 
Therefore, a PEF is an exponential family with polynomial sufficient statistics $t(x)=(x, x^2,\ldots,x^D)$. 
However, the log-normalizer $F_D(\theta)=\log\int \exp(\theta^\top t(x))\dx$ of a $D$-order 
PEF is not available in closed-form.
Nevertheless, the KL of two mixtures $m(x)$ and $m'(x)$ can be {\em theoretically} approximated 
closely by a Bregman divergence: 
$\KL(m(x):m'(x)) \simeq \KL(p(x;\theta):p(x;\theta')) = B_{F_D}(\theta':\theta)$, 
where $\theta$ and $\theta'$ 
are the natural parameters of the PEF family $\{p(x;\theta)\}$ approximating $m(x)$ and $m'(x)$, respectively 
(i.e., $m(x)\simeq p(x;\theta)$ and $m'(x)\simeq p(x;\theta')$).

\item Consider two finite mixtures $m(x)=\sum_{i=1}^k w_ip_i(x)$ and $m'(x)=\sum_{j=1}^{k'} w_j'p_j'(x)$ 
of  $k$ and $k'$ components (possibly with heterogeneous components $p_i(x)$'s and $p_j'(x)$'s),
respectively.
In information geometry, a mixture family is the set of convex combination of fixed\footnote{Thus in statistics,
a mixture is understood as a convex combination of parametric components while in information geometry
a mixture family is the set of convex combination of fixed components.} component densities.
Let us consider the mixture families $\{g(x;(w,w'))\}$ generated by the $D=k+k'$ fixed components 
$p_1(x),\ldots, p_k(x), p_1'(x),\ldots, p_{k'}'(x)$: 
$$\left\{
g(x;(w,w'))
=
\sum_{i=1}^k w_ip_i(x) +\sum_{j=1}^{k'}w_{j}'p_{j}'(x)
\ :\
\sum_{i=1}^k w_i +\sum_{j=1}^{k'} w_{j}' =1
\right\}$$
We can approximate arbitrarily finely mixture $m(x)$ for any $\epsilon>0$ by 
$g(x;\alpha)\simeq (1-\epsilon)m(x)+\epsilon m'(x)$ with $\alpha=((1-\epsilon)w,\epsilon w')$ (so that $\sum_{i=1}^{k+k'} \alpha_i=1$) 
and
$m'(x)\simeq  g(x;\alpha')= \epsilon m(x)+(1-\epsilon) m'(x)$ with $\alpha'=(\epsilon w,(1-\epsilon) w')$ (and $\sum_{i=1}^{k+k'} \alpha_i'=1$).
Therefore $\KL(m(x):m'(x))\simeq \KL(g(x;\alpha):g(x;\alpha')) = B_{F^*}(\alpha:\alpha')$,
where $F^*(\alpha)=\int g(x;\alpha)\log g(x;\alpha)\dx$ is the Shannon information (negative Shannon entropy) for the composite mixture family.
Interestingly, this Shannon information can be arbitrarily closely approximated when considering isotropic Gaussians~\cite{HGMM-2016}.
Notice that the convex conjugate $F(\theta)$ of the continuous Shannon neg-entropy $F^*(\eta)$ 
is the log-sum-exp function on the  inverse soft map.
\end{itemize}

\appendix

\section{The Kullback-Leibler divergence of mixture models is not analytic~\protect\cite{KLnotanalytic-2004}\label{sec:KLnonanalytic}}

Ideally, we aim at getting a finite length closed-form formula to 
compute the KL divergence of mixture models.
But this is provably mathematically intractable because of the log-sum 
term in the integral, as we shall prove below.
Analytic expressions encompass closed-form formula and include special 
functions (e.g., Gamma function)
but do not allow to use limits nor integrals.
An analytic function $f(x)$ is a $C^\infty$ function  (infinitely 
differentiable) such that at any point $x_0$
its $k$-order Taylor series $T_k(x)=\sum_{i=0}^k 
\frac{f^{(i)}(x_0)}{i!}(x-x_0)^i$
converges to $f(x)$: $\lim_{k\rightarrow \infty} T_k(x) = f(x)$ for $x$ 
belonging to a neighborhood
$N_r(x_0)=\{ x : |x-x_0|\leq r\}$ of $x_0$ where
$r$ is called the radius of convergence.
The analytic property of a function is equivalent to the condition that 
for each $k\in\mathbb{N}$,
there exists a constant $c$ such that
$\left| \frac{\mathrm{d}^k f}{\mathrm{d}x^k}(x) \right| \leq c^{k+1} k!$.

To prove that the KL of mixtures is not analytic (hence does not admit a 
closed-form formula), we
shall adapt the proof reported in~\cite{KLnotanalytic-2004}
(in Japanese\footnote{We thank Professor Aoyagi for sending us his 
paper~\cite{KLnotanalytic-2004}.}).
We shall prove that $\KL(p:q)$ is not analytic for univariate mixtures
of densities $p(x)=G(x;0,1)$ and $q(x;w)=(1-w)G(x;0,1)+w G(x;1,1)$ for 
$w\in (0,1)$, where
$G(x;\mu,\sigma)=\frac{1}{\sqrt{2\pi}\sigma}\exp(-\frac{(x-\mu)^2}{2\sigma^2})$ 
is the density of a univariate Gaussian of mean $\mu$ and standard 
deviation $\sigma$. Let $D(w)=\KL(p(x):q(x;w))$ denote the
divergence between these two mixtures ($p$ has a single component and 
$q$ has two components).

We have $\log\frac{p(x)}{q(x;w)}=-\log(1+w (e^{x-\frac{1}{2}}-1))$, and
$$
\frac{\mathrm{d}^k D}{\mathrm{d}w^k} = \frac{(-1)^k}{k}\int 
p(x)(e^{x-\frac{1}{2}}-1) \dx.
$$
Let $x_0$ be the root of the equation $e^{x-\frac{1}{2}}-1= 
e^{\frac{x}{2}}$ so
that for $x\ge x_0$, we have $e^{x-\frac{1}{2}}-1 \ge e^{\frac{x}{2}}$.
It follows that:

$$
\left| \frac{\mathrm{d}^k D}{\mathrm{d}w^k}  \right| \geq \frac{1}{k} 
\int_{x_0}^\infty p(x) e^{\frac{kx}{2}} \dx
=  \frac{1}{k} e^{\frac{k^2}{8}} A_k
$$
with $A_k=\int_{x_0}^\infty \frac{1}{\sqrt{2\pi}} 
\exp(-\frac{x-\frac{k}{2}}{2}) \dx$.
When $k\rightarrow\infty$, we have $A_k\rightarrow 1$.
Consider $k_0\in\mathbb{N}$ such that $A_{k_0}>0.9$.
Then the radius of convergence $r$ is such that:

$$
\frac{1}{r}\geq \lim_{k\rightarrow\infty} \left(\frac{1}{k k!} 0.9 
\exp\left(\frac{k^2}{8}\right) \right)^{\frac{1}{k}} =\infty.
$$

Thus the convergence radius is $r=0$, and therefore the KL divergence is 
not an analytic function of the parameter $w$.
The KL of mixtures is an example of a non-analytic smooth function.
(Notice that the absolute value is not analytic at $0$.)

\section{Closed-form formula for the Kullback-Leibler divergence between scaled and truncated exponential families\label{sec:appendix}}

When computing approximation bounds for the KL divergence between two mixtures $m(x)$ and $m'(x)$, we end up
with the task of computing $\int_{\calD} w_a p_a(x)\log \frac{w_b' p_b'(x)}{w_c' p_c'(x)}\dx$ where $\calD\subseteq\calX$ is a subset of the full support $\calX$.
We report a generic formula for computing these formula when the mixture (scaled and truncated) components belong to the same exponential family~\cite{ef-flashcard-2009}.
An exponential family has canonical log-density written as $l(x;\theta)=\log p(x;\theta)=\theta^\top t(x)-F(\theta)+k(x)$, where $t(x)$ denotes the sufficient satistics, $F(\theta)$ the log-normalizer (also called cumulant function or partition function), and $k(x)$ an auxiliary carrier term.

Let $\KL(w_1p_1:w_2p_2:w_3p_3)=\int_\calX w_1p_1(x)\log\frac{w_2p_2(x)}{w_3p_3(x)}\dx=\cH(w_1p_1:w_3p_3)-\cH(w_1p_1:w_2p_2)$.
Since it is a difference of two cross-entropies, we get
  for three distributions belonging to the same exponential family~\cite{cH-EF-2010} the following formula: 
$$
\KL(w_1p_1:w_2p_2:w_3p_3)=w_1\log\frac{w_2}{w_3}+w_1(F(\theta_3)-F(\theta_2)-(\theta_3-\theta_2)^\top\nabla F(\theta_1)).
$$

Furthermore, when the support is restricted, say to support range $\calD\subseteq\calX$, let $m_\calD(\theta)=\int_\calD p(x;\theta)\dx$ denote the mass
and $\tilde{p(x;\theta)}=\frac{p(x;\theta)}{m_\calD(\theta)}$ the normalized distribution. 
Then we have:

$$
\int_\calD w_1p_1(x)\log\frac{w_2p_2(x)}{w_3p_3(x)}\dx=
m_\calD(\theta_1)(\KL(w_1\tilde{p}_1:w_2\tilde{p}_2:w_3\tilde{p}_3)) - \log \frac{w_2 m_\calD(\theta_3)}{w_3 m_\calD(\theta_2)}.
$$

When $F_\calD(\theta)=F(\theta)-\log m_\calD(\theta)$ is strictly convex and differentiable 
then $\tilde{p(x;\theta)}$ is an exponential family and the closed-form formula follows straightforwardly.
Otherwise, we still get a closed-form but need more derivations.
For univariate distributions, we write $\calD=(a,b)$ and $m_\calD(\theta)=\int_a^b p(x;\theta)\dx=P_\theta(b)-P_\theta(a)$
where $P_\theta(a)=\int^a p(x;\theta)\dx$ denotes the cumulative distribution function.

The usual formula for truncated and scaled Kullback-Leibler divergence is:
\begin{equation}
\KL_\calD(w p(x;\theta) : w' p(x;\theta')) =
w m_\calD(\theta) \left(\log\frac{w}{w'}+B_F(\theta':\theta)\right) + w {(\theta'-\theta)}^\top {\nabla m_\calD(\theta)}\label{eq:cfgeneric},
\end{equation}
where $B_F(\theta':\theta)$ is a Bregman divergence~\cite{bregmankmeans-2005}:
$$
B_F(\theta':\theta)=F(\theta')-F(\theta)-(\theta'-\theta)^\top \nabla F(\theta).
$$

This formula extends the classic formula~\cite{bregmankmeans-2005} for full regular exponential families (by setting $w=w'=1$
 and $m_\calD(\theta)=1$ with $\nabla m_\calD(\theta)=0$).

Similar formula are available for the cross-entropy and entropy of exponential families~\cite{cH-EF-2010}.

\bibliographystyle{plain}
\bibliography{KLGMM-LSE-BIB2}

\end{document}